\documentclass{midl} 

\usepackage{wrapfig}
\usepackage{caption}
\usepackage{multicol}

\jmlrproceedings{}{}
\jmlrpages{}
\jmlryear{}
\usepackage{hyperref}

\title[CellNet]{CellNet - Localizing Cells using Sparse and Noisy Point Annotations}

\midlauthor{
\Name{Benjamin Eckhardt\nametag{$^{1,3}$}} \Email{benjamin.eckhardt@stud.uni-goettingen.de}
\AND
\Name{Stuart Fawke\nametag{$^{2}$}} \Email{sf24@sanger.ac.uk}
\AND
\Name{Andrew Curtis\nametag{$^{2}$}} \Email{ac64@sanger.ac.uk}
\AND
\Name{Bo Fussing\nametag{$^{2}$}} \Email{bf15@sanger.ac.uk}
\AND
\Name{Dmytro Fishman\midljointauthortext{Equal Supervision}\nametag{$^{3}$}} \Email{dmytro.fishman@ut.ee}
\AND
\Name{Constantin Pape\midlotherjointauthor\nametag{$^{1}$}} \Email{constantin.pape@informatik.uni-goettingen.de}\\
\addr $^{1}$ University of Göttingen, 
\addr $^{2}$ Wellcome Sanger Institute, 
\addr $^{3}$ University of Tartu
}

\begin{document}

\maketitle

\begin{abstract}
Counting living cells is an important step in many biological research workflows. Our
collaborators at the Wellcome Sanger Institute study vital genes in humans
via large scale saturation genome editing screening, which requires repeatedly counting cells a great
number of times. Computer Vision based automation is crucial for high throughput and resource efficiency.
In this work, we develop a regression-based deep learning computer vision algorithm to
detect and count cells in phase-contrast microscopy images. To reduce annotation effort,
which in practice often becomes a bottleneck, we focus on counting cells only using sparse
point annotations, which are fast and easy to acquire. By comparison to state-of-the-art
0-shot methods, we show that regression-based counting is a promising alternative in low data regimes.
Through developing methods to automatically count living cells in microscopy images, we
contribute to valuable research on the human genome. Code \footnote{\hyperlink{https://github.com/beijn/cellnet}{https://github.com/beijn/cellnet}}.
\end{abstract}


\section{Introduction}
Automated cell counting is essential for high-throughput genome research, such as mapping vital genes by quantifying surviving cells after targeted mutations. Current semi-manual workflows using devices like the Countess 3 FL (ThermoFisher) are costly, time-consuming, and prone to inaccuracies. Approaches for counting cells can be categorized as counting by detection and counting by regression \cite{lempitsky2010learning}.
On the detection side, Kirillov et al.'s Segment Anything (SAM) \cite{kirillov2023segment} achieves zero-shot instance segmentation across diverse domains. Recently, Pachitariu et al. introduced CellposeSAM \cite{pachitariu2025cellpose}, combining Cellpose and SAM for state-of-the-art zero-shot cellular segmentation without manual prompts or domain-specific fine-tuning.
On the regression side, Lempitsky et al. \cite{lempitsky2010learning} introduced counting objects via density maps constructed as superpositions of Gaussian distributions around point annotations, requiring only minimal ground truth. Xie et al. \cite{xie2018microscopy} applied fully convolutional networks to predict such density maps.  
This idea has been extended to sparse point annotations and 3D data \cite{malin2023automated} as well as to objects of varying size \cite{jeremias2025stamped}.
We build on these ideas with CellNet, a U-Net \cite{ronneberger2015u} with a ResNet \cite{he2016deep} backbone, and demonstrate that this lightweight, point-annotation-based approach can be promising to replace 0-shot detection-based methods while minimizing annotation effort and computational cost.

\section{Methods}
\begin{wrapfigure}{r}{0.38\textwidth}
    \centering
    \includegraphics[width=0.32\linewidth]{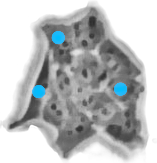}
    \includegraphics[width=0.32\linewidth]{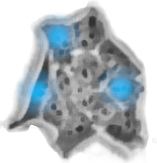}
    \includegraphics[width=0.32\linewidth]{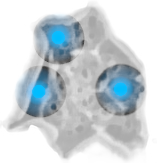}
    \caption{Construction of cell indicator ground truth from sparse point annotations of cells. (Left) Sparse point annotations of cell centers (blue), 2D projection. (Middle) Blurring with Gaussian distributions. (Right) Exclusion of unannotated regions from training (bright shade).}
    \label{fig:method}
\end{wrapfigure}

\subsection{Model}
We develop CellNet, a fully convolutional neural network (CNN), to predict an object density map, similar to the work by Xie et al. \cite{xie2018microscopy}. The local maxima in the density map correspond to the locations of the objects, in our case cells. The number of objects in an image patch is determined by the integral of its density map \cite{lempitsky2010learning}.
The ground truth density map is constructed as the sum of Gaussian probability distributions centered around each point annotation with constant standard deviation \cite{lempitsky2010learning, xie2018microscopy}, as shown in Figure \ref{fig:method}. We chose the standard deviation as high as possible without visually merging the distributions of the two pairwise closest annotations. Areas without annotations are excluded from training by multiplying the pixel-wise loss with a mask similar to \cite{malin2023automated}. We train using MSE+BCE loss until convergence.

\subsection{Datasets, Metrics and Baselines}
We evaluate our approach on a dataset of 3 sparsely point-annotated phase-contrast images dubbed \emph{Sanger23}, two of which we use for training and one for testing. Secondly we apply our approach to the LivECell dataset \cite{edlund2021livecell} by training on the centers of mass of the segmentation masks from the 2\% train split. As a baseline for counting by detection, we employ CellposeSAM \cite{pachitariu2025cellpose}. To assess counting performance, we report the symmetric mean absolute percentage error (sMAPE) and mean absolute error (MAE), which quantify the accuracy of predicted cell counts relative to the ground truth. Additionally, we use the F1 score to evaluate the localization performance of detected cells based on local maxima detection and Hungarian matching.
Furthermore, we compared our predicted counts to a Countess 3 FL Automated Cell Counter (ThermoFisher) \footnote{https://www.thermofisher.com/tw/zt/home/life-science/cell-analysis/cell-analysis-instruments/automated-cell-counters/models/countess-3-fl.html [thermofisher.com, retrieved 13.04.2026]} using a dataset of 840 phase-contrast images from 105 flasks across 6 cell lines and up to 8 time points without annotations. There we averaged over 8 images per flask and scaled by the flask area over the microscope's field of vision.

\section{Results}
\vspace{-0.5em}

\begin{wrapfigure}{r}{0.6\textwidth}
  \vspace{-\intextsep}
  \centering
  \begin{tabular}{lllll}
  \bfseries Dataset & \bfseries Model & \bfseries sMAPE & \bfseries MAE & \bfseries F1\\
  LiveCell & Cellpose & 7.94\% & 57.80 & 0.02 \\
   & CellNet & 30.50\% &  148.06 &  0.02 \\ 
  Sanger23 & Cellpose & 19.09\% & 452.00 & 0.05 \\
   
  & CellNet* & 51.04\% &1186.12 & 0.01 \\
  \end{tabular}
  \captionof{table}{Results comparing CellNet to CellposeSAM on LiveCell and Sanger23 datasets. * = evaluation using only one image.}
  \label{tab:results}

  \vspace{0.5em}

  \includegraphics[width=0.43\linewidth]{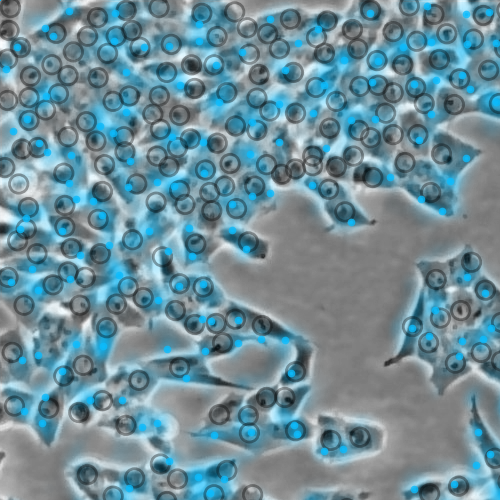}
  \hfill
  \includegraphics[width=0.55\linewidth]{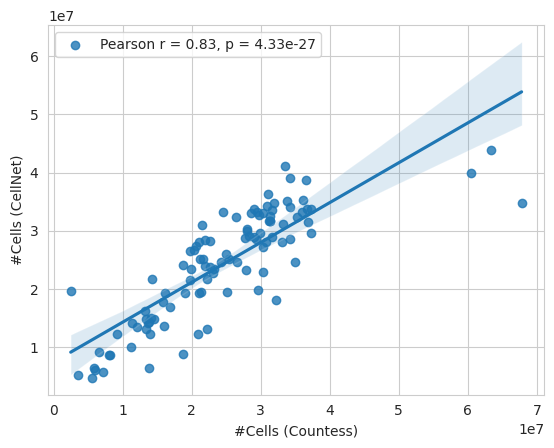}
  \captionof{figure}{(Left) Test image overlayed with predicted density map (blue shade), local maxima used for localization and F1-scoring (blue dots), and ground-truth annotation points (black circles); (Right) Comparison with Countess 3 FL.}
  \label{fig:figure}
  \vspace{-\intextsep}
\end{wrapfigure}

Table \ref{tab:results} summarizes the counting performance of CellposeSAM and CellNet on the LiveCell and Sanger23 datasets. On LiveCell, CellposeSAM achieves an sMAPE of 7.94\% and MAE of 57.80, substantially outperforming the current version of CellNet (sMAPE 30.50\%, MAE 148.06). On the 3-image dataset Sanger23, CellposeSAM leads as well with an sMAPE of 19.09\% and MAE of 452.00, while CellNet, evaluated on only a single image, yields an sMAPE of 51.04\% and MAE of 1186.12. The low F1 scores across both datasets and methods probably indicate a methodological flaw in the matching procedure. Figure \ref{fig:figure} (left) illustrates a representative CellNet prediction, showing the regressed density map and local maxima overlayed on a test image alongside ground-truth annotations. While predicted cell densities coincide with the overall concentration of cells, exact localization of cell centers is still challenging, possibly due to label noise. Figure \ref{fig:figure} (right) compares our results to the Countess 3 FL device on a larger dataset of 105 flasks with an 8-average of image counts each. We report a strong correlation ($r=0.83, p=4.22\times10^{-27}$) which seems to degrade slightly at very high cell densities. A possible reason for this might be that the model is not generalizing well to images of much higher density than the training images; possibly due to overlap and cell deformation. The spread might be partly due to the random sampling effects of counting only a small subset of the whole flask area.

\vspace{-1.2em}

\section{Conclusion}
\vspace{-0.5em}
We present a regression-based counting method, \emph{CellNet} using few sparse point annotations. Our preliminary results are outperformed by the SOTA 0-shot generalization capacities of CellposeSAM. However, counting by regression remains promising due to its very low data need, the sufficiency of approximate locations, and architectural light-weightedness. Particularly, a counting model with performance comparable to specialized wet lab techniques could be trained with only two partially annotated images. Future directions include adopting a pretrained backbone similar to CellposeSAM, improving the loss to reflect integrated counts directly, and improving the density map generation for high confluency.
Thereby, we hope to increase the accuracy of this approach in order to match or outperform 0-shot methods in challenging settings such as bad data quality due to label efficiency.

\vspace{-3em}
\midlacknowledgments{I am grateful for Dmytro Fishman, for being an invested supervisor, caring for my personal wellbeing
and academic success during and beyond this work. I thank Stuart Fawke, our numinous collaborator
from the Welcome Sanger Institute, for this opportunity to make a contribution to meaningful research,
and for kindly spending his time and energy to provide us data and explanations; in what felt like
a prosperous collaboration transcending occasional misunderstandings. I thank Constantin Pape for
being a strong source of ideas and guidance, whenever his time allowed, and I asked early enough.
I thank the University of Tartu Institute of Computer Science for providing me access to research-accelerating hardware. I especially thank Anne-Kathrin Schultz for being a very awesome and invested
study advisor, making possible my globe spanning academic trajectories.}

\bibliography{bibliography.bib}

\begin{thebibliography}{9}
\providecommand{\natexlab}[1]{#1}
\providecommand{\url}[1]{\texttt{#1}}
\expandafter\ifx\csname urlstyle\endcsname\relax
  \providecommand{\doi}[1]{doi: #1}\else
  \providecommand{\doi}{doi: \begingroup \urlstyle{rm}\Url}\fi

\bibitem[Edlund et~al.(2021)Edlund, Jackson, Khalid, Bevan, Dale, Dengel,
  Ahmed, Trygg, and Sj{\"o}gren]{edlund2021livecell}
Christoffer Edlund, Timothy~R Jackson, Nabeel Khalid, Nicola Bevan, Timothy
  Dale, Andreas Dengel, Sheraz Ahmed, Johan Trygg, and Rickard Sj{\"o}gren.
\newblock Livecell—a large-scale dataset for label-free live cell
  segmentation.
\newblock \emph{Nature methods}, 18\penalty0 (9):\penalty0 1038--1045, 2021.

\bibitem[He et~al.(2016)He, Zhang, Ren, and Sun]{he2016deep}
Kaiming He, Xiangyu Zhang, Shaoqing Ren, and Jian Sun.
\newblock Deep residual learning for image recognition.
\newblock In \emph{Proceedings of the IEEE conference on computer vision and
  pattern recognition}, pages 770--778, 2016.

\bibitem[Jeremias and Pape(2025)]{jeremias2025stamped}
Julia Jeremias and Constantin Pape.
\newblock Stamped counting for biomedical images.
\newblock \emph{Methods in Microscopy}, 2\penalty0 (2):\penalty0 203--213,
  2025.

\bibitem[Kirillov et~al.(2023)Kirillov, Mintun, Ravi, Mao, Rolland, Gustafson,
  Xiao, Whitehead, Berg, Lo, et~al.]{kirillov2023segment}
Alexander Kirillov, Eric Mintun, Nikhila Ravi, Hanzi Mao, Chloe Rolland, Laura
  Gustafson, Tete Xiao, Spencer Whitehead, Alexander~C Berg, Wan-Yen Lo, et~al.
\newblock Segment anything.
\newblock \emph{arXiv preprint arXiv:2304.02643}, 2023.

\bibitem[Lempitsky and Zisserman(2010)]{lempitsky2010learning}
Victor Lempitsky and Andrew Zisserman.
\newblock Learning to count objects in images.
\newblock \emph{Advances in neural information processing systems}, 23, 2010.

\bibitem[Malin-Mayor et~al.(2023)Malin-Mayor, Hirsch, Guignard, McDole, Wan,
  Lemon, Kainmueller, Keller, Preibisch, and Funke]{malin2023automated}
Caroline Malin-Mayor, Peter Hirsch, Leo Guignard, Katie McDole, Yinan Wan,
  William~C Lemon, Dagmar Kainmueller, Philipp~J Keller, Stephan Preibisch, and
  Jan Funke.
\newblock Automated reconstruction of whole-embryo cell lineages by learning
  from sparse annotations.
\newblock \emph{Nature biotechnology}, 41\penalty0 (1):\penalty0 44--49, 2023.

\bibitem[Pachitariu et~al.(2025)Pachitariu, Rariden, and
  Stringer]{pachitariu2025cellpose}
Marius Pachitariu, Michael Rariden, and Carsen Stringer.
\newblock Cellpose-sam: superhuman generalization for cellular segmentation.
\newblock \emph{BioRxiv}, pages 2025--04, 2025.

\bibitem[Ronneberger et~al.(2015)Ronneberger, Fischer, and
  Brox]{ronneberger2015u}
Olaf Ronneberger, Philipp Fischer, and Thomas Brox.
\newblock U-net: Convolutional networks for biomedical image segmentation.
\newblock In \emph{Medical image computing and computer-assisted
  intervention--MICCAI 2015: 18th international conference, Munich, Germany,
  October 5-9, 2015, proceedings, part III 18}, pages 234--241. Springer, 2015.

\bibitem[Xie et~al.(2018)Xie, Noble, and Zisserman]{xie2018microscopy}
Weidi Xie, J~Alison Noble, and Andrew Zisserman.
\newblock Microscopy cell counting and detection with fully convolutional
  regression networks.
\newblock \emph{Computer methods in biomechanics and biomedical engineering:
  Imaging \& Visualization}, 6\penalty0 (3):\penalty0 283--292, 2018.

\end{thebibliography}
\vspace{-100em}
\end{document}